\pdfoutput=1
\documentclass[11pt]{article}

\PassOptionsToPackage{table}{xcolor}

\usepackage[preprint]{acl}

\usepackage{times}       % 这行确保使用Times Roman字体
\usepackage{latexsym}
\usepackage{makecell}

\usepackage[T1]{fontenc}
\usepackage[utf8]{inputenc}

\usepackage{microtype}

\usepackage{inconsolata}
\usepackage{subcaption}

\usepackage{tcolorbox}
\tcbuselibrary{skins,breakable}

\usepackage{array}        % 用于高级表格格式
\usepackage{colortbl}     % 用于表格背景色
\definecolor{simvbgcolor}{rgb}{0.95,0.95,0.85}

\newtcolorbox{promptbox}[1][]{
  enhanced,
  colback=white,
  colframe=black,
  arc=7pt,
  boxrule=1pt,
  fonttitle=\bfseries,
  coltitle=black,
  attach boxed title to top left={yshift=-\tcboxedtitleheight/2},
  boxed title style={colback=white, colframe=white},
  width=\textwidth,
  #1
}

\newtcolorbox{stepbox}[2][]{
  enhanced,
  colback=white,
  colframe=black,
  arc=7pt,
  boxrule=1pt,
  title={\textbf{Step #2}},
  fonttitle=\bfseries,
  coltitle=black,
  attach boxed title to top left={yshift=-\tcboxedtitleheight/2},
  boxed title style={colback=white, colframe=white},
  width=\textwidth,
  #1
}

\usepackage{graphicx}

\usepackage{amsmath}
\usepackage{amssymb}

\usepackage{booktabs}    % 更专业的表格线
\usepackage{multirow}    % 支持表格中的多行单元格

\title{ValueSim: Generating Backstories to Model Individual Value Systems}

\author{\textbf{Bangde Du\textsuperscript{1}, \ Ziyi Ye\textsuperscript{1}, \ Zhijing Wu\textsuperscript{2}, \ Monika Jankowska\textsuperscript{3}, \ Shuqi Zhu\textsuperscript{1}, \ Qingyao Ai\textsuperscript{1}}, \\ 
\textbf{Yujia Zhou\textsuperscript{1}, \ Yiqun Liu\textsuperscript{1}\thanks{$^{\ast}$Corresponding author}}  \\
  \textsuperscript{1}Department of Computer Science and Technology, Tsinghua University, Beijing, China \\
  \textsuperscript{2}Beijing Institute of Technology, Beijing, China \\
  \textsuperscript{3}Rice University, Houston, United States \\
  \texttt{dbd23@mails.tsinghua.edu.cn, yeziyi1998@gmail.com, wuzhijing.joyce@gmail.com} \\
  \texttt{monika.a.jankowska@gmail.com, zsq19991106@gmail.com, aiqy@tsinghua.edu.cn} \\
  \texttt{zhouyujia@ruc.edu.cn, yiqunliu@tsinghua.edu.cn} \\
  }

\begin{document}
\maketitle

\begin{abstract}
As Large Language Models (LLMs) continue to exhibit increasingly human-like capabilities, aligning them with human values has become critically important. Contemporary advanced techniques, such as prompt learning and reinforcement learning, are being deployed to better align LLMs with human values. However, while these approaches address broad ethical considerations and helpfulness, they rarely focus on simulating individualized human value systems.
To address this gap, we present \textbf{ValueSim}, a framework that simulates individual values through the generation of personal backstories reflecting past experiences and demographic information. \textbf{ValueSim} converts structured individual data into narrative backstories and employs a multi-module architecture inspired by the Cognitive-Affective Personality System to simulate individual values based on these narratives.
Testing \textbf{ValueSim} on a self-constructed benchmark derived from the World Values Survey demonstrates an improvement in top-1 accuracy by over 10\% compared to retrieval-augmented generation methods. Further analysis reveals that performance enhances as additional user interaction history becomes available, indicating the model's ability to refine its persona simulation capabilities over time.
\end{abstract}

% 引入各个章节
\section{Introduction}
% 1. 大模型可以模拟人类完成很多任务，助力社科和心理学研究等。有利于被试规模大、或者有伦理问题的社科研究
% 2. 价值观的定义，不同价值观的人会有不同的...大模型需要在不同场景下模拟个体的价值观来完成模拟人类的任务，现有的工作还没做过
% 3. 价值观的形成是咋样的->backstory具有表征价值观的能力
% 4. 我们提出并构建一个框架，使用backstory表征人类价值观的方法来模拟人，结合了ABC理论构建multi-system。
% 5. 我们在wvs上构建bench并测试。实验结果表明，该方法可以更好地模拟人的价值观，尤其是（效果比较好的子任务），并且价值观模拟效果随着对这个用户的交互数据的采集不断提升，backstory的幻觉同时存在正向和负向的效应。
% 6. summarize the contribution.

% Large language models' (LLMs) remarkable progress in simulating human behaviors and cognition~\cite{brown2020language,touvron2023llama}, created unprecedented opportunities to build agents with distinct personalities reflecting real people or constituting more realistic synthetic personas. 
Large language models (LLMs) have demonstrated strong capabilities to simulate humans' behaviors and cognition~\cite{brown2020language,touvron2023llama}.
Such human simulation created unprecedented opportunities to build agents with distinct personalities reflecting real people or constituting realistic synthetic personas.
These LLM-empowered agents have multiple uses, encompassing social science and behavioral research simulations ~\cite{park2023generative,aher2023using}, and serving as personalized assistants, tutors, or partners. 
To create such agents, existing studies have tried to infuse LLMs with the knowledge and understanding of the individuals whom those agents are expected to mimic or interact with.
For example, \citet{wang2023humanoid} proposes to train an LLM with reinforcement learning to mimic the behaviors and preferences of various population groups.
\citet{tu2023characterchat, wang2023incharacter, wang2023rolellm} adopt pre-generated persona-based information
profiles to prompt an LLM to act as different characters.   

% In this study, we propose novel approaches to providing LLMs with information about people with the goal of building LLM agents that reflect individuals' characteristics more accurately. % 第一段这个写法不太像是计算机论文

% Currently, the research on AI agents tends to focus on simulating populations rather than individuals~\cite{argyle2023out,santurkar2023whose}, which is crucial for understanding and mitigating biases in LLM agents' responses \cite{boelaert2025machine,wang2025large}. 
As LLMs exhibit such human-like and simulation capabilities, the challenge of aligning them with human values becomes critically important.
Conventionally, LLMs are aligned with general human values such as helpfulness, honesty, harmlessness, and fairness, with techniques such as system prompt design~\cite{guo2024review} and reinforcement learning~\cite{christiano2017deep,liu2020learning,ye2025learning}. 
% which serve as foundational guardrails for safe and broadly acceptable interactions.
% A variety of techniques, such as system prompt design~\cite{guo2024review}, reinforcement learning~\cite{christiano2017deep,liu2020learning,ye2025learning}, have been developed to facilitate such value alignment in LLMs.
% Aligning with general human values enables large models to better serve humanity. 
% While those value alignments serve as foundational guardrails for safe and broadly acceptable interactions, while crucial, represents only one facet of the challenge. 
% However, true human values can differ significantly across individuals, cultures, and contexts.
Alignment with these general values provides crucial foundational guardrails for safe and broadly acceptable interactions.
However, true human values can differ significantly across individuals, cultures, and contexts, presenting further complexities.
% Existing focus on value alignment, 
It is equally important for LLMs to develop the capacity to simulate individual values that are more fine-grained and varied.

% it remains challenging for makes it more challenging for them to reflect the diverse values of individuals.
% This research also indicates that the LLM agents' responses show limited variations across populations and individuals. 
% To address this problem, our research focuses on simulating the answers of specific individuals rather than the members of certain populations in order to build agents that reflect the diversity of human beliefs and behaviors more accurately. 

This research uses information about personal values and beliefs to better simulate an individual's personality and preferences. 
Values—the enduring beliefs that guide attitudes and behaviors~\cite{schwartz2012overview}—fundamentally shape how individuals perceive and interact with their world. 
People with differing value systems respond distinctively to identical situations, making value representation essential for accurate human simulation in various contexts, from conducting behavioral research to building personal assistants. 

\begin{figure*}[t]
  \centering
  \includegraphics[width=\textwidth]{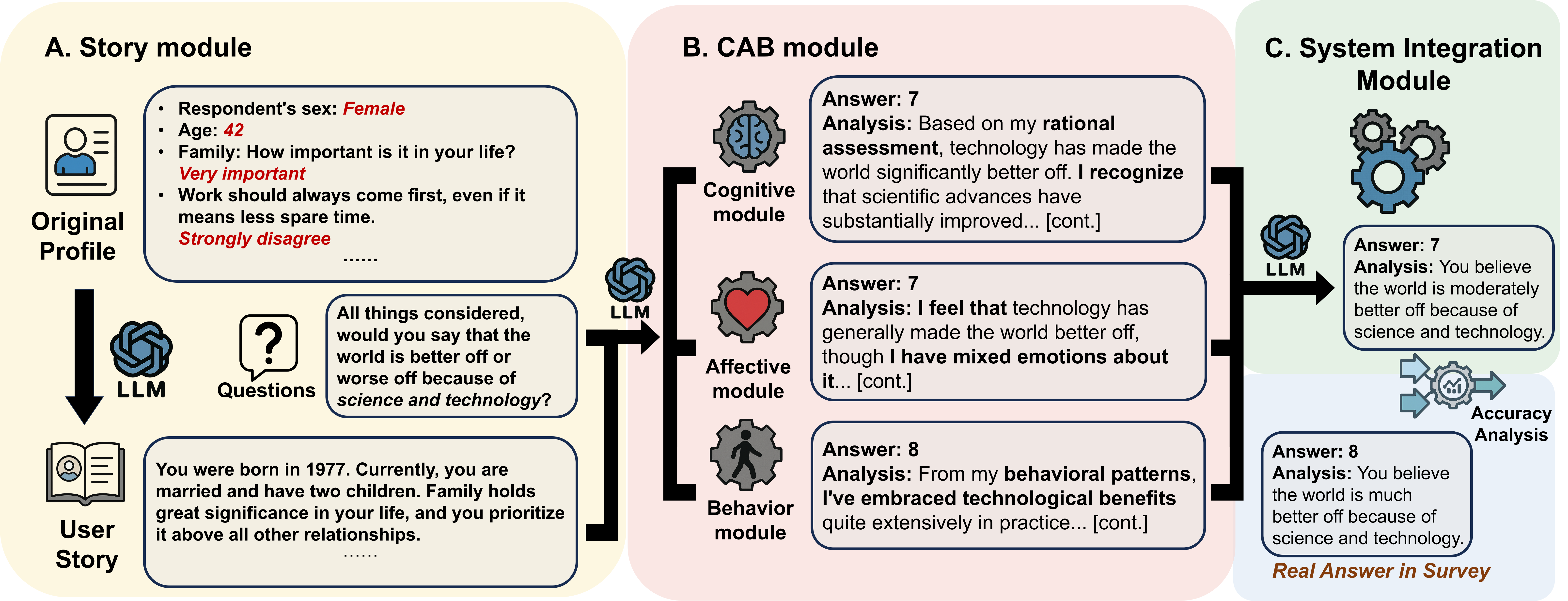}
  \caption{The complete process of simulating human value responses. For each individual, the profile is first converted into a story through the story module, then, using the cognitive-affective-behavior module and after system integration, the individual's response to a specific question is simulated.}
  \label{fig:procedure}
\end{figure*}

To bridge this gap, we investigate the simulation of individual values with LLMs. 
Motivated by the fact that the formation and expression of human values stem from complex personal histories and experiences, we propose a simple yet effective method, ValueSim~(Simulating Individual Values by Backstory Generation), to prompt an LLM with individual backstories to simulate human behavior.
While previous studies have used personas' profiles to create LLM agents that would represent certain populations \cite{moon2024virtual, wang2025large,park2024generative}, backstories encapsulate critical life events, cultural contexts, and social influences that collectively shape an individual's value system~\cite{mcadams2001psychology}. 
% we leverage the rich nature of backstories to move beyond demographic archetypes towards a more authentic simulation of individual values.
% Here, we propose a new approach to provide LLMs with the information about simulated individuals in a more efficient manner.
Specifically, ValueSim leverages backstory-based value representation combined with a multi-system architecture grounded in the Cognitive-Affective System Theory of Personality~(CAPS)~\cite{mischel1995cognitive}. 
% 这么写的话读者可能会不能理解query
As shown in Figure~\ref{fig:procedure}, ValueSim consists of three modules.
(1)~First, a story module prompts an LLM to write a backstory with information about an individual encompassing their demographic profile and response to a series of questions related to their value. % their可作为单数
The performance of the backstory is evaluated using questions that are independent of those used to generate the backstory, yet are responded to by the same individual.
(2)~Second, a Cognitive-Affective-Behavioral~(CAB) module motivated by the CAPS is devised to understand the backstories from different aspects and generate multiple candidate responses to simulate the individual.
This approach mirrors the multifaceted nature of human value-driven responses, which rarely emerge from isolated cognitive or emotional processes~\cite{loewenstein2003role}. 
(3)~Finally, a system integration module is used to integrate outputs from the CAB modules and generate the response that is most likely to mirror the real answer responded by the individual.  
% For each query, our system simultaneously models responses through cognitive, emotional, and behavioral perspectives, with results integrated through a dedicated synthesis module. 

% While fully grasping LLMs' ``thinking'' processes is impossible, the growing body of research indicates that LLMs' performance benefits from providing them with more structured "thinking" patterns and opportunities for self-evaluation \cite{shinn2023reflexion, wang2022self, zhou2022least}. 
% Hence, we can draw inspiration from behavioral and social science theories to design architectures that encourage LLMs to "think" longer and in a more organized and self-reflective manner. 
% ValueSim exemplifies this approach. 

To evaluate our framework, we constructed a comprehensive benchmark based on the World Values Survey dataset~\cite{haerpfer2022world}, comprising 97,220 individuals with 290 distinct demographic and opinion attributes. 
Our experimental results demonstrate that ValueSim significantly outperforms existing methods in simulating human value-based responses, including those with users' full information and the Retrieval-augmented Generation~(RAG) system that retrieves historical individual information for each response generation.
ValueSim achieves the most performance gain, particularly in domains related to happiness perception and social value judgments. 
Furthermore, we observe that simulation accuracy improves progressively with increased interaction data from target individuals, showing the system's capacity for personalization refinement over more interactions~\cite{hwang2023aligning}.

Our contributions include:
\begin{itemize}
    \item We present a framework that leverages backstories, derived from raw profile data, to effectively prompt Large Language Models (LLMs) in representing individual values.
    \item To simulate individual responses, we propose a modular, personality theory-inspired method that captures the cognitive, affective, and behavioral dimensions of human cognitive processes.
    \item We empirically show that the proposed method outperforms a series of existing prompting-based methods to represent human value and demonstrate that the alignment accuracy improves with increased individual information.
    % A backstory processing approach that transforms raw profile data into organized narratives representing human values, enhancing the simulation's fidelity.
    % \item Empirical evidence demonstrating that while simulation accuracy improves with increased interaction data, the rate of improvement diminishes significantly after a certain threshold.
\end{itemize}

% Through ValueSim, we establish a new paradigm for human simulation with LLMs, which leverages efficient organization of information about an individual in the form of "backstories" and combines it with multidimensional processing to capture the richness of human responses.

\section{Related Work}

\subsection{User-based Alignment in Large Language Models}
Recent research has explored various approaches to align language models with user characteristics and preferences. \citet{sun2024random} introduced "Random Silicon Sampling" to simulate human sub-populations using group-level demographic information, while~\citet{moon2024virtual} developed "Anthology," employing personal backstories to create virtual personas for improved alignment with survey responses.
\citet{hwang2023aligning} found that demographics and ideologies alone are insufficient predictors of user opinions, demonstrating that incorporating relevant past opinions improves accuracy in predicting responses to survey questions.

In personalized approaches, Personalized-RLHF~\citep{li2024personalized} captures individual preferences through a lightweight user model trained jointly with the LLM. For specific applications,~\citet{xiang2024simuser} created SimUser for simulating user interactions with mobile applications. These works represent important steps toward user simulation, yet they typically rely on narrowly defined contexts. Our work extends these efforts by utilizing data describing human values and preferences to capture an individual's personality more accurately. Furthermore, we implement a structured cognitive-affective-behavioral framework and demonstrate effectiveness across diverse questions with verifiable ground truth responses.

\subsection{Multi-System Approaches to Human Decision-Making} 
Contemporary psychological research recognizes that human cognition, emotion, and behavior arise from parallel, interacting neural systems~\cite{pessoa2008relationship, kahneman2011thinking}. 
For instance, the Cognitive-Affective System Theory of Personality (CAPS)~\cite{mischel1995cognitive} that describes personality as a dynamic network of cognitive-affective units—including encodings, expectancies, emotions, and behavioral competencies—that activate concurrently in response to situational features. Metacognition research further reveals how the brain coordinates these parallel systems, resolving conflicts and integrating diverse processing streams into the final decisions \cite{fleming2012neural}. 

These theoretical foundations suggest that authentic simulation of human decision-making requires modeling the distinct contributions of cognitive, affective, and behavioral processes, along with the integrative mechanisms that harmonize their parallel operations.

Even though LLMs process information in different ways than humans, the recent research indicates that they are increasingly capable of mimicking complex cognitive processes \cite{kosinski2024evaluating, xie2024can}, raising questions that CAPS and other psychological theories can be used to build technical frameworks enabling better information processing by the LLMs in the context of human simulations.

\subsection{Value System Modeling and WVS Studies}
Scholars like \cite{hofstede1983national,schwartz2012toward, inglehart2006mapping} developed distinct frameworks for classifying and analyzing values on individual and community levels. 
Those frameworks were used to examine the relationship between values and beliefs held by individuals and pro-environmental behaviors \cite{primc2021does}, responses to corporate social responsibility initiatives \cite{rosario2014values}, and the broadly defined national culture characteristics \cite{beugelsdijk2018dimensions}. 

WVS occupies an important place in human value studies due to its extensive geographic and coverage (including beliefs and attitudes about religion, economy, science and technology), the number of respondents, and data accessibility \footnote{https://www.worldvaluessurvey.org/wvs.jsp}. 
For this reason, the WVS data is frequently used to better understand topics such as people's perception of well-being \cite{fleche2012exploring}, support for democracy \cite{ariely2011can}, or the relationship between trust and health \cite{jen2010trustful}. 
WVS data, are also used in LLM-related research. However, the LLM-related studies using WVS data focus on studying groups of people \cite{alkhamissi-etal-2025-llm}, rather than individuals, with a view to using LLM agents for participation in social surveys \cite{boelaert2025machine}.

\section{Methdology}

We present a novel framework for human value response simulation that addresses two fundamental challenges: representing complex user profiles and capturing the multidimensional nature of human decision-making. 
Our approach consists of two primary components. 
First, a Story Processing Module transforms survey responses into coherent narrative representations, enabling more effective integration of user information. 
Second, a Multi-Module Simulation Framework decomposes the response generation process into parallel cognitive, affective, and behavioral dimensions to more accurately represent human decision-making. 
This modular approach enables more nuanced modeling of personality-specific response patterns than monolithic prompt-based methods. 
We complement these core components with a data configuration to enable rigorous evaluation of simulation performance across diverse user profiles and survey sections.

\subsection{Task Definition}
In this work, we address the challenge of simulating human responses to value-oriented questions based on personal profiles. Formally, given a user's profile information collected from previous survey responses, our task is to predict how that individual would respond to new value-related questions not contained in their profile.

Let $P = {(q_1, a_1), (q_2, a_2), ..., (q_n, a_n)}$ represent a user's profile, where each pair $(q_i, a_i)$ consists of a survey question and the user's corresponding answer. The profile information includes both multiple-choice questions with their selected options and fill-in-the-blank questions with the provided answers. In our dataset, profiles contain up to 232 question-answer pairs used for training.

For a new question $q_{new}$ not present in $P$, the task is to predict the user's response $\hat{a}_{new}$, which is typically a selection from a set of predefined options for multiple-choice questions (the predominant format in our value survey questions).

This task presents two significant challenges. First, the user profiles contain approximately 290 pieces of information (with 232 used for training), making it difficult for models to identify which profile elements are relevant to a particular prediction. Second, the overall complexity of human cognitive, emotional, and behavioral processes, as well as individual differences between people in this area, make the simulation of specific people's answers extremely difficult.

We evaluate performance using two metrics: Accuracy, which measures the percentage of correctly predicted responses for multiple-choice questions, and Mean Absolute Error (MAE), which quantifies the average deviation between predicted and actual responses when answers can be ordinally ranked. These metrics are calculated for each simulated individual and then averaged across all individuals to evaluate the overall framework performance.

\subsection{Story Module}
To facilitate accurate user-level alignment of the LLM-generated responses with complex profiles, we introduce a story processing module that transforms survey responses into coherent, narrative-based representations.

\subsubsection{Motivation of Story Module}

Direct incorporation of full people's profiles into prompt contexts presents several challenges: 
(1) \textbf{Context Length Limitations}: Even with modern LLMs' expanded context windows, including hundreds of question-answer pairs remains inefficient; 
(2) \textbf{Information Integration}: Raw question-answer formats impede the model's ability to form holistic representations of individual personalities; 
(3) \textbf{Retrieval Limitations}: Conventional retrieval-augmented generation (RAG) approaches that select only subsets of profile information for each query sacrifice the holistic understanding of an individual's personality pattern.

As demonstrated in our experiments (Section~5.1), both direct input and retrieval-based approaches fail to achieve optimal alignment with actual human responses, either overwhelming the model with disjointed information or providing incomplete information.

\subsubsection{Narrative Transformation Technique}
We developed a specialized narrative transformation technique that converts structured survey data into coherent backstories while preserving information integrity. 
This technique leverages cognitive principles of narrative processing, which suggest that humans (and by extension, LLMs trained on human text) better comprehend and retain information presented in story format compared to disconnected factual statements.

Our transformation framework operates through a two-phase approach:
\begin{enumerate}
\item \textbf{Thematic Organization}: Survey responses are reorganized according to conceptual relatedness (demographic attributes, value systems, political orientations, etc.), establishing coherent narrative threads
\item \textbf{Narrative Integration}: These thematically organized elements are then woven into a continuous second-person narrative that maintains the accuracy of the content while providing natural transitions and logical flow between concepts
\end{enumerate}

The transformation framework enforces strict information preservation constraints, ensuring that specific values, numerical responses, and unique identifiers remain unaltered during transformation. 
Meanwhile, the narrative structure facilitates holistic comprehension by explicitly connecting related beliefs and attributes that might otherwise remain implicit in raw survey data.

This approach yields personalized narratives that: (1) maintain complete fidelity to the original 232 profile elements; (2) reduce cognitive load for the LLMs through coherent organization; and (3) make it easier for the downstream model to identify relevant personality patterns during response simulation.

\subsection{Multi-Module Simulation Framework}
Building on the narrative representation provided by the story processing module, we introduce a simulation framework that helps capture the multifaceted nature of human decision-making. 
Our approach draws inspiration from established theories in cognitive psychology and neuroscience, particularly the Cognitive-Affective Personality System (CAPS) theory~\cite{mischel1995cognitive}, which conceptualizes personality as a dynamic network of cognitive-affective units that activate in situation-specific patterns.

\subsubsection{Cognitive-Affective-Behavioral Architecture}
 We decompose the simulation process into three parallel processing modules corresponding to the primary dimensions identified in cognitive neuroscience research:

\begin{enumerate}
\item \textbf{Cognitive Module}: Simulates information processing, reasoning patterns, belief structures, and analytical tendencies that influence decision-making
\item \textbf{Affective Module}: Models emotional responses, value alignments, motivational states, and affective reactions to potential outcomes
\item \textbf{Behavioral Module}: Captures action tendencies, implementation patterns, contextual influences, and behavioral histories
\end{enumerate}

This tripartite architecture is grounded in CAPS theory's classification of personality units~\cite{mischel1995cognitive}, which delineates encodings and expectancies (cognitive), affects (emotional), and competencies and self-regulatory plans (behavioral) as parallel processing components. Our design mirrors the parallel activation dynamics described in neuropsychological models of decision-making \citep{pessoa2008relationship}, where cognitive, affective, and behavioral systems operate concurrently before integration.

\subsubsection{Module Implementation}
Each module employs specialized prompting to simulate its respective psychological domain:

\begin{enumerate}
\item \textbf{Cognitive Module}: Processes profile information with emphasis on belief structures, information gathering preferences, reasoning approaches, worldview framing, and weighting factors in analytical decision-making. 
%This implementation aligns with dual-process theories of reasoning \citep{evans2008dual} and models of belief-based inference.
\item \textbf{Affective Module}: Analyzes profile information with focus on affective patterns, emotional regulation styles, values, motivational states, and identity-based emotional responses. 
%This approach is informed by appraisal theories of emotion \citep{scherer2009emotions} and value-based decision models.
\item \textbf{Behavioral Module}: Examines profile information through the lens of behavioral tendencies, environmental influences, capability constraints, experiential learning, and implementation patterns. 
%This design draws from behavioral economics and situated action theories.
\end{enumerate}

Each module generates both a predicted response option and a detailed analysis of the reasoning process from its specialized perspective. This parallel processing approach captures the specialized contribution of each psychological system while avoiding the limitations of monolithic simulation methods.

% \subsubsection{System Integration Module}
% Introduction介绍时分了三个模块，建议这里也按照三个模块来介绍，改成一级子章节
\subsection{System Integration Module}
The System Integration Module synthesizes outputs from the three specialized modules into a coherent final response. Rather than simply averaging numerical predictions, this module analyzes the detailed reasoning provided in each analysis to identify patterns of alignment, conflict, and contextual dominance. This qualitative integration may better capture how individuals reconcile potentially conflicting cognitive, affective, and behavioral tendencies—a critical aspect of authentic decision-making overlooked by simplistic aggregation methods.

In our ablation studies (Section~5.2), we compare this integration approach with a baseline that simply averages the predictions from the three modules, showing the contribution of structured integration to prediction accuracy.
 
% 增加一个section进行任务定义
\section{Experimental Setup}
\subsection{Data Configuration} % 这一章放在methods里不太合适，建议挪到新加的第四章Experimental Settings，或者直接挪到第四章里，章标题改名为Experiments and Results
\subsubsection{Dataset Selection}
For the value response simulation task, we required a dataset with sufficient scale, coverage, and profile richness to effectively develop and evaluate our framework.
We selected the World Values Survey (WVS) Wave 7 dataset, collected between 2017 and 2022, encompassing user information from 66 distinct countries. 
This dataset provides a globally diverse sample with comprehensive value profiles spanning various cultural contexts.

Each user in the WVS dataset responded to 290 questions, covering both demographic information (age, gender, education level, income, etc.) and multidimensional value information (political views, religious beliefs, social attitudes, environmental concerns, etc.). This rich profile data enables the construction of detailed user representations necessary for nuanced personality simulation.

\subsubsection{Training-Testing Partition}
We implemented a 4:1 training-testing split to enable rigorous evaluation of the framework's performance. The training partition contains 80\% of user profiles, which are used to construct the narrative representations for personality simulation. The testing partition, comprising the remaining 20\% of user profiles, is reserved for evaluation purposes, where we compare the framework's predicted responses against actual user responses on the same questions.

To ensure robust and unbiased evaluation, we conducted five-fold cross-validation, systematically rotating which portion of the data serves as the test set. This approach guarantees that our performance metrics reflect the framework's generalization capability across different subsets of the population rather than potential artifacts of a specific data split.

\subsection{Language Models}
We evaluated SimVBG using four diverse language models:

GPT-3.5-Turbo~\cite{brown2020language}: OpenAI's commercial model with instruction-tuning and RLHF optimization.
Llama-3.1-8B~\cite{touvron2023llama}: Meta AI's open-source model with 8 billion parameters and strong multilingual capabilities.
Qwen-2.5-7B~\cite{bai2023qwen}: Alibaba Cloud's model with 7 billion parameters featuring an extended context window and specialized training on reasoning tasks.
DeepSeek-V3~\cite{liu2024deepseek}: A recent foundation model optimized for dialogue coherence and knowledge representation.

For reproducibility, we set the temperature parameter to zero across all models and maintained identical prompting frameworks for all experimental conditions.
% 数据集构建，baseline方法实现，一些细节（比如rag用的检索模型）的实现，大模型的调用方法等细节
\section{Experiments and Results}

\subsection{Main Results}

\begin{table*}[t]
\centering
\small
\setlength{\tabcolsep}{0.8pt}
\renewcommand{\arraystretch}{1.1}
\definecolor{ValueSimcolor}{rgb}{1.0,0.95,0.8}
\caption{Mean Absolute Error (MAE) performance comparison across different value dimensions and methods (lower is better). Column headers represent different value dimensions from the World Values Survey: Core Values (Core), Happiness \& Well-being (Hap.), Trust, Economic Integrity (Econ.Int), Security, Technology (Tech), Moral \& Religious (Mo.\&Rel.), Political Engagement (Pol.Eng), and Demographics (Demo). Complete descriptions of these value dimensions are provided in Appendix~F.}
\begin{tabular}{p{1.23cm}l>{\centering\arraybackslash}p{1.3cm}>{\centering\arraybackslash}p{1.3cm}>{\centering\arraybackslash}p{1.3cm}>{\centering\arraybackslash}p{1.3cm}>{\centering\arraybackslash}p{1.3cm}>{\centering\arraybackslash}p{1.3cm}>{\centering\arraybackslash}p{1.3cm}>{\centering\arraybackslash}p{1.3cm}>{\centering\arraybackslash}p{1.3cm}>{\centering\arraybackslash}p{1.3cm}}
\toprule
\textbf{Model} & \textbf{Method} & \makecell[c]{Core} & \makecell[c]{Hap.} & \makecell[c]{Trust} & \makecell[c]{Econ.Int} & \makecell[c]{Security} & \makecell[c]{Tech} & \makecell[c]{Mo.\&Rel.} & \makecell[c]{Pol.Eng} & \makecell[c]{Demo} & \makecell[c]{\textbf{Overall}} \\
\midrule
\multirow{3}{=}{GPT-3.5 -Turbo} & Full Info & 0.597 & 0.438 & 0.352 & 0.326 & 0.460 & 0.425 & 0.378 & 0.489 & 0.556 & 0.465 \\        
  & RAG & 0.363 & 0.423 & 0.247 & 0.274 & 0.302 & 0.265 & 0.240 & 0.371 & 0.483 & 0.336 \\
 & \cellcolor{ValueSimcolor}\textit{\textbf{ValueSim}} & \cellcolor{ValueSimcolor}\textbf{0.299} & \cellcolor{ValueSimcolor}\textbf{0.167} & \cellcolor{ValueSimcolor}\textbf{0.231} & \cellcolor{ValueSimcolor}\textbf{0.239} & \cellcolor{ValueSimcolor}\textbf{0.280} & \cellcolor{ValueSimcolor}\textbf{0.246} & \cellcolor{ValueSimcolor}\textbf{0.188} & \cellcolor{ValueSimcolor}\textbf{0.265} & \cellcolor{ValueSimcolor}\textbf{0.344} & \cellcolor{ValueSimcolor}\textbf{0.260} \\
\midrule
\multirow{3}{=}{Llama- 3.1-8B} & Full Info & 0.543 & 0.660 & 0.324 & 0.326 & 0.439 & 0.411 & 0.408 & 0.434 & 0.592 & 0.452 \\
  & RAG & 0.453 & 0.563 & 0.297 & 0.307 & 0.381 & \textbf{0.326} & 0.308 & 0.400 & 0.536 & 0.390 \\
& \cellcolor{ValueSimcolor}\textit{\textbf{ValueSim}} & \cellcolor{ValueSimcolor}\textbf{0.400} & \cellcolor{ValueSimcolor}\textbf{0.222} & \cellcolor{ValueSimcolor}\textbf{0.226} & \cellcolor{ValueSimcolor}\textbf{0.302} & \cellcolor{ValueSimcolor}\textbf{0.346} & \cellcolor{ValueSimcolor}0.343 & \cellcolor{ValueSimcolor}\textbf{0.250} & \cellcolor{ValueSimcolor}\textbf{0.308} & \cellcolor{ValueSimcolor}\textbf{0.326} & \cellcolor{ValueSimcolor}\textbf{0.308} \\
\midrule
\multirow{3}{=}{Qwen- 2.5-7B} & Full Info & 0.669 & 0.315 & 0.367 & 0.325 & 0.383 & 0.374 & 0.492 & 0.488 & 0.592 & 0.477 \\
  & RAG & 0.777 & 0.598 & 0.359 & 0.328 & 0.550 & 0.653 & 0.492 & 0.557 & 0.584 & 0.544 \\
& \cellcolor{ValueSimcolor}\textit{\textbf{ValueSim}} & \cellcolor{ValueSimcolor}\textbf{0.341} & \cellcolor{ValueSimcolor}\textbf{0.201} & \cellcolor{ValueSimcolor}\textbf{0.219} & \cellcolor{ValueSimcolor}\textbf{0.253} & \cellcolor{ValueSimcolor}\textbf{0.320} & \cellcolor{ValueSimcolor}\textbf{0.331} & \cellcolor{ValueSimcolor}\textbf{0.236} & \cellcolor{ValueSimcolor}\textbf{0.293} & \cellcolor{ValueSimcolor}\textbf{0.360} & \cellcolor{ValueSimcolor}\textbf{0.288} \\
\midrule
\multirow{3}{=}{DeepSeek -V3} & Full Info & 0.365 & 0.611 & 0.258 & 0.382 & 0.341 & \textbf{0.233} & 0.287 & 0.325 & 0.530 & 0.355 \\ 
  & RAG & 0.360 & 0.444 & 0.236 & \textbf{0.253} & 0.275 & 0.260 & 0.229 & 0.280 & 0.466 & 0.304 \\
  & \cellcolor{ValueSimcolor}\textit{\textbf{ValueSim}} & \cellcolor{ValueSimcolor}\textbf{0.306} & \cellcolor{ValueSimcolor}\textbf{0.168} & \cellcolor{ValueSimcolor}\textbf{0.210} & \cellcolor{ValueSimcolor}0.270 & \cellcolor{ValueSimcolor}\textbf{0.267} & \cellcolor{ValueSimcolor}0.251 & \cellcolor{ValueSimcolor}\textbf{0.197} & \cellcolor{ValueSimcolor}\textbf{0.257} & \cellcolor{ValueSimcolor}\textbf{0.338} & \cellcolor{ValueSimcolor}\textbf{0.257} \\
\midrule
\makecell[c]{--} & Chance & 0.550 & 0.486 & 0.497 & 0.414 & 0.539 & 0.393 & 0.474 & 0.463 & 0.610 & 0.507 \\
\bottomrule
\end{tabular}
\label{tab:mae_results}
\end{table*}

Our experiments evaluate ValueSim against baseline approaches on value response alignment tasks using four LLMs: Llama-3.1-8B, Qwen-2.5-7B, GPT-3.5-Turbo, and DeepSeek-V3. 
%For Llama-3.1-8B and Qwen-2.5-7B, we conducted comprehensive evaluations with 100 test samples each, while for GPT-3.5-Turbo and DeepSeek-V3, we used a smaller set of 10 samples due to computational constraints.
With our limited computational resources, we tested all models on 100 samples (simulating 100 different users). 
Our framework is designed to support the simulation and testing of all 97,220 users in our dataset.

We compared our framework against two baseline methods: (1) Full Info, which directly inputs the complete profile information, and (2) RAG, which selectively provides the most relevant profile information. 
The Full Info method inputs all 232 survey questions and responses from a user's profile alongside the target question to generate LLM predictions. 
The RAG method employs text-embedding-ada-002 to create embeddings for each question and profile entry, then uses cosine similarity to identify the 3 most relevant profile entries for each test question.

ValueSim consistently outperforms both baseline methods across the tested LLMs. 
As shown in Table~\ref{tab:mae_results}, our approach demonstrates substantial improvements in both accuracy and Mean Absolute Error (MAE) for the Llama-3.1-8B and Qwen-2.5-7B models, with similar patterns for GPT-3.5-Turbo and DeepSeek-V3.
Paired t-tests confirm that these improvements are statistically significant when comparing ValueSim against both Full Info ($p < 0.05$) and RAG ($p < 0.05$) methods overall.
These results indicate that our simulation approach captures underlying preference patterns more effectively than methods that either use complete profile information or select information based solely on semantic similarity. The advantage appears to derive from ValueSim's structured simulation of human decision processes rather than from model-specific characteristics.

\subsection{Ablation Studies}
To validate the contribution of each component in our ValueSim framework, we conducted a series of ablation experiments across all four LLM models.

%\begin{table*}[t]
%\centering
%\small
%\begin{tabular}{lccccccc}
%\toprule
%\multirow{2}{*}{\textbf{LLM Model}} & \multicolumn{2}{c}%{\textbf{ValueSim (ours)}} & \multicolumn{2}{c}{\textbf{w/o CAB %Module}} & \multicolumn{2}{c}{\textbf{w/o Story Module}} \\
%\cmidrule(lr){2-3} \cmidrule(lr){4-5} \cmidrule(lr){6-7}
%& Accuracy & MAE & Accuracy & MAE & Accuracy & MAE \\
%\midrule
%GPT-3.5-Turbo (10 sets) & 0.548 & 0.218 & 0.489 & 0.333 & 0.435 & %0.323 \\
%Llama-3.1-8B (100 sets) & 0.494 & 0.260 & 0.486 & 0.311 & 0.517 & %0.270 \\
%Qwen-2.5-7B (100 sets) & 0.516 & 0.231 & 0.427 & 0.400 & 0.475 & %0.278 \\
%GPT-4o-mini (10 sets) & 0.593 & 0.205 & 0.544 & 0.260 & 0.517 & %0.267 \\
%\bottomrule
%\end{tabular}
%\caption{Ablation study results comparing our full ValueSim model %against variants without the Cognitive-Affective-Behavior (CAB) %module and without the Story module across different LLM models. %Metrics reported are Accuracy and Mean Absolute Error (MAE).}
%\label{tab:ablation}
%\end{table*}

\begin{table*}[t]
\centering
\small
\setlength{\tabcolsep}{4pt}
\renewcommand{\arraystretch}{1.1}
\definecolor{ValueSimcolor}{rgb}{1.0,0.95,0.8}
\caption{Ablation study on ValueSim across different language models in terms of MAE~(lower is better). }
\begin{tabular}{lcccc}
\toprule
\textbf{Setting} & \textbf{GPT-3.5-Turbo} & \textbf{Llama-3.1-8B} & \textbf{Qwen-2.5-7B} & \textbf{Deepseek-V3} \\
\midrule
\rowcolor{ValueSimcolor} ValueSim & \cellcolor{ValueSimcolor}\textbf{0.264} & \cellcolor{ValueSimcolor}\textbf{0.310} & \cellcolor{ValueSimcolor}\textbf{0.271} & \cellcolor{ValueSimcolor}\textbf{0.244} \\
ValueSim w/o CAB module & 0.352 & 0.322 & 0.341 & 0.321 \\
ValueSim w/o story module & 0.304 & 0.312 & 0.308 & 0.256 \\
\bottomrule
\end{tabular}
\label{tab:ablation_mae}
\end{table*}

\subsubsection{Story module contribution analysis}
First, we examined the impact of the story generation module by replacing it with the original profile information while maintaining the parallel question-answering structure of subsequent modules. Results consistently demonstrated that the story module significantly enhances framework performance across all tested LLMs.

\subsubsection{Three-module parallel structure contribution analysis}
Next, we assessed the cognitive-affective-behavioral (CAB) module as a whole by removing it entirely and using only the generated story for direct response prediction. This ablation revealed substantial performance degradation across all models, confirming that the structured simulation of mental processes provided by the CAB module is crucial for accurate value alignment.

We further investigated the individual contributions of each unit within the CAB module by selectively removing one unit while keeping the others intact. These fine-grained ablations showed that each component plays a distinct and necessary role in the overall framework

\subsection{Impact of the User Profile Scale}
To investigate how the amount of profile information influences simulation accuracy, we conducted an incremental profile scale experiment using our ValueSim framework. 
Given that user profiles contain substantial information (232 training data points per user), we sought to understand the relationship between profile comprehensiveness and simulation performance.

% \subsubsection{Experimental Setup for Incremental Profile Testing}
% 这里建议不要进一步划分子章节了，否则results里有个醒目的子章节标题讲setup比较怪，尤其这个实验并不是主要的重点内容
We maintained the same testing conditions as our main experiment, using the same test-train split (20\% as test questions, 80\% as training questions) from one fold of our cross-validation setup. 
For each user, we varied the amount of profile information provided to the model by randomly sampling from their available training data points. 
We tested five configurations: 0, 58, 116, 174, and 232 profile items, with the sampling increment of 58 chosen to provide sufficient granularity while maintaining experimental feasibility. 
For each configuration, we randomly sampled the specified number of profile items for each user to ensure unbiased comparison.

\begin{figure}[t]
    \centering
    \includegraphics[width=\columnwidth]{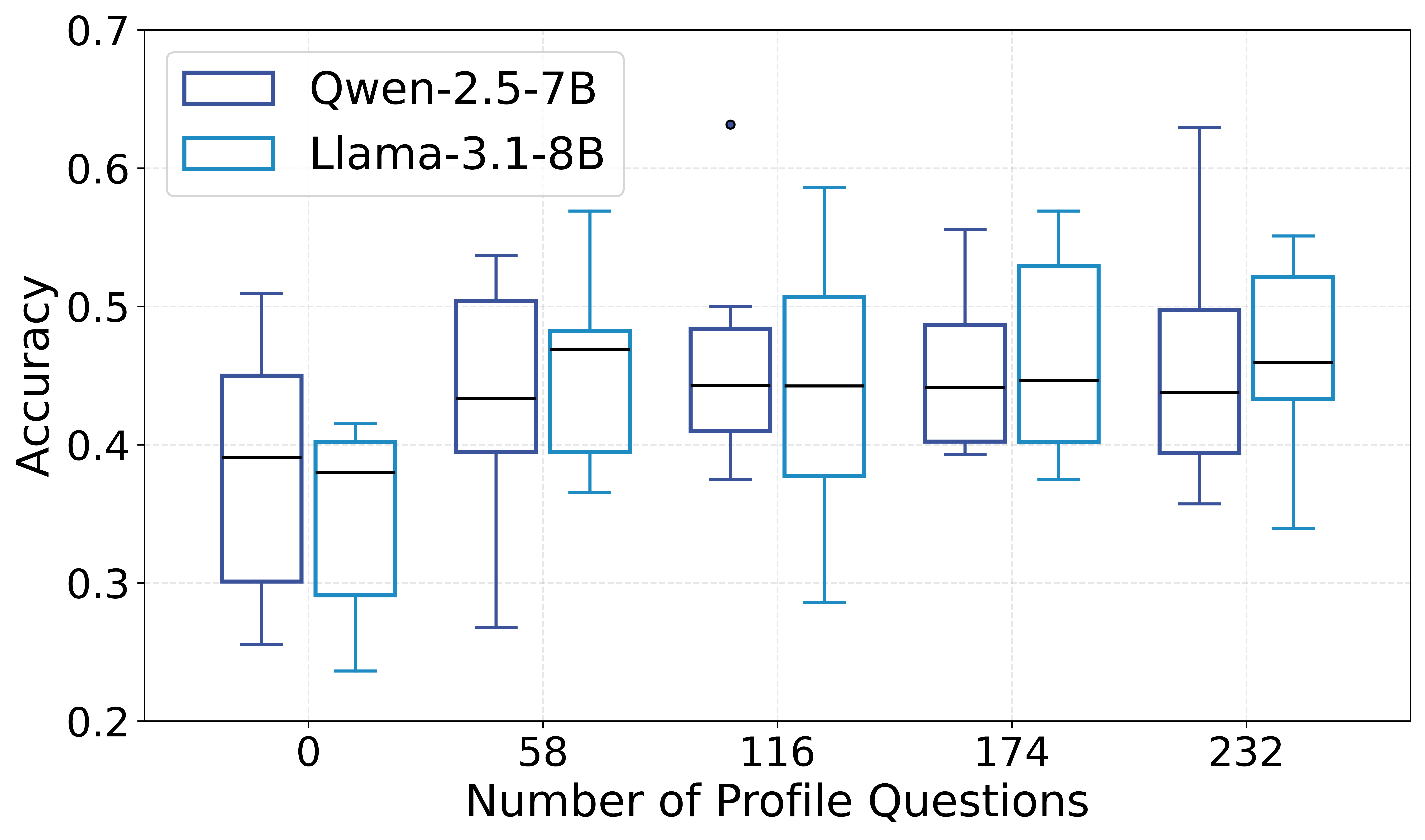}
    \caption{Impact of profile information scale on simulation accuracy.}
    \label{fig:profile_scale_accuracy}
\end{figure}

\begin{figure}[t]
    \centering
    \includegraphics[width=\columnwidth]{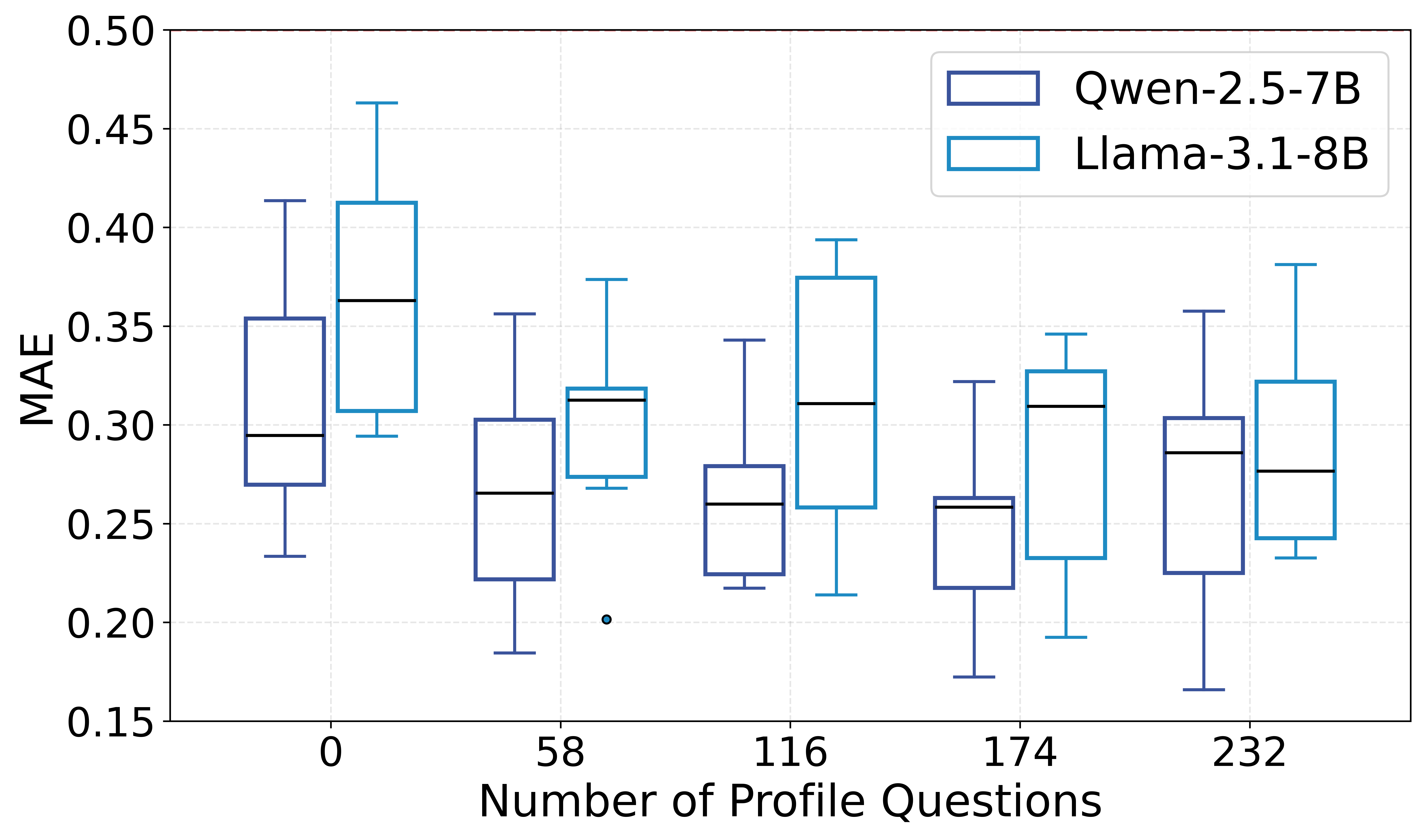}
    \caption{Impact of profile information scale on simulation error (lower is better).}
    \label{fig:profile_scale_mae}
\end{figure}

% \subsubsection{Performance Trends with Increasing Profile Information}
Our findings demonstrate that increasing the amount of profile information consistently improves simulation performance across all four model architectures. 
As shown in Figure~\ref{fig:profile_scale_accuracy} and Figure~\ref{fig:profile_scale_mae}, both accuracy and MAE metrics improve as more profile information becomes available.

Notably, we observed that the largest improvements occur in the early stages of profile expansion (from 0 to approximately 100 items). 
Beyond this threshold, while performance continues to improve, the rate of improvement diminishes, indicating diminishing returns when profile information exceeds approximately 100 items.

This pattern reveals that in value response alignment tasks, expanding user profiles to include several dozen data points provides substantial benefits for simulation accuracy. 
However, the incremental benefit of additional profile information decreases as profiles become more comprehensive. 
This finding has implications for designing profile-based simulation systems, suggesting an optimal balance between profile comprehensiveness, participants' effort and privacy, and computational efficiency.

%\subsection{illusion analysis}

%此章一般和前一章放一起
\section{Conclusion}
We presented ValueSim, a framework for simulating human value responses by backstories that combines narrative generation with a structured cognitive-affective-behavioral processing. 
Our experiments across multiple base LLMs showed that ValueSim consistently outperforms methods using either complete user information or a retrieval-augmented generation method. 
Ablation studies confirmed that both the story generation module and the cognitive-affective-behavioral module contribute substantially to the framework's effectiveness.

The success of ValueSim suggests that simulating human responses benefits from a modular approach inspired by psychological processes. 
By generating backstories that contextualize abstract values and then processing these through parallel cognitive, affective, and behavioral pathways, ValueSim more accurately learned from the raw information included in individuals' profiles to make value judgments more aligned with the individuals. 
Further research could explore more sophisticated architectural designs and leverage advanced post-training techniques for LLMs to enhance their capabilities in producing and processing backstories to reflect human values.

This work has potential applications in personalized AI systems and social science research. 
For personalized services, frameworks like ValueSim could enable systems that adapt to individual preferences without requiring extensive data collection. 
In social science, such simulations could help explore societal dynamics by modeling interactions between individuals with different value systems.

\clearpage

% Limitations部分(必需)
\section*{Limitations}
\subsection*{Technical Limitations}

Our study has several limitations that should be addressed in future work.

First, our evaluation focused exclusively on value-related questions from the World Values Survey, leaving open the question of whether SimVBG would be effective for simulating responses in other domains. Further research could incorporate other types of data into the SimVBG framework.

Second, creating backstories that are naturalistic but also include all the important information about an individual constitutes a significant challenge. The process of translating the raw information about a person into a coherent backstory that would mirror the way this person describes their experiences may result in hallucinations or misportrayal of certain elements of a person's profile. Simultaneously, making the narrative more realistic requires a certain level of transformation of raw data, in a manner that may better convey a person's emotion, preferences, or communication style. Further research on how to achieve the right balance between the two is necessary.

Third, due to computational constraints, we tested our approach on only 100 simulated users, though our framework supports all 97,220 users in our dataset. More comprehensive testing could reveal additional insights about performance across diverse demographic groups and value systems.

Finally, while our cognitive-affective-behavioral module draws inspiration from psychological theories, it remains a simplification of actual human mental processes. 
Future work could explore whether more sophisticated psychological models can further enhance LLM simulation of human behavior.

\subsection*{Ethical Concerns and Societal Implications}

We used a publicly available dataset about people's values, hence being non-invasive from the perspective of individuals' privacy. In general, however, simulating individuals' opinions and behaviors poses a unique challenge of achieving the balance between the usefulness of simulation for the particular goals and respecting the person's privacy and autonomy.

Personalized agents that can either behave in a more realistic "human" way or just understand preferences of specific individuals have the potential for use in a variety of settings, e.g., providing more individualized education resources, assisting in various jobs, serving as companions for elderly people, and participating in some social and behavioral research. While using LLM agents for these purposes may have a significant positive societal impact, we need to indicate certain ethical challenges surrounding this endeavor. 

The attempts to simulate an individual's opinions and behaviors may encourage a tendency to collect extensive and sensitive data about the individuals. Hence, we need a legal and ethical framework protecting individuals from undue infringements of their privacy and allowing individuals to have influence over how their data is used.

Furthermore, personalized agents may misrepresent certain individuals or even act in a manipulative way. Hence, the actual use of such agents requires technical and institutional safeguards that would consider holistically the circumstances of the agents' usage, especially the purpose of such usage, the levels and nature of the relevant risks, as well as the need to respect individuals' privacy and autonomy.

% 可选的伦理考虑部分
%\section*{Ethical Considerations}
%\input{ethics}

% 致谢(可选，仅在最终版本)
%\section*{Acknowledgments}
%\input{acknowledgments}

% 参考文献
% 使用ACL风格引用
\bibliography{custom}
% 如果只使用自定义文献，可以用：
% \bibliography{custom}

% 附录(如有需要)
\appendix
\clearpage
\onecolumn
\appendix

\section*{Appendix}
\label{sec:appendix}

% Overview of all appendix sections with hyperlinks and increased spacing
\noindent\textbf{Appendix \hyperref[app:main_experiments]{A}} provides the accuracy results of our main experiments compared with two baselines.

\vspace{0.5cm}

\noindent\textbf{Appendix \hyperref[app:ValueSim_values]{B}} presents the radar chart showing the effectiveness of the ValueSim framework across different types of values.

\vspace{0.5cm}

\noindent\textbf{Appendix \hyperref[app:ValueSim_prompt]{C}} details the prompt flow process of the ValueSim framework.

\vspace{0.5cm}

\noindent\textbf{Appendix \hyperref[app:backstory_examples]{D}} provides real examples of backstories generated in our experiments.

\vspace{0.5cm}

\noindent\textbf{Appendix \hyperref[app:baseline_prompts]{E}} includes the prompts used for the two baselines: Origin Full and RAG.

\vspace{0.5cm}

\noindent\textbf{Appendix \hyperref[app:wvs_dataset]{F}} provides an introduction to the World Value Survey (WVS) dataset used in our study.

\vspace{1cm}

% Appendix A
\clearpage
\section{Accuracy Results of Experiments}
\label{app:main_experiments}

This section presents comprehensive accuracy results from our main experiments, providing a direct comparison between our proposed approach and two baseline methods, as well as the accuracy results from our ablation studies.
While the key findings are discussed in the main text, it is worth noting that our ValueSim approach achieves accuracy comparable to Mean Absolute Error (MAE) metrics, demonstrating its effectiveness in realistic user simulation.

\begin{table*}[h]
\centering
\small
\setlength{\tabcolsep}{2pt}
\renewcommand{\arraystretch}{1.1}
\definecolor{ValueSimcolor}{rgb}{1.0,0.95,0.8}
\begin{tabular}{llcccccccccc}
\toprule
\textbf{Model} & \textbf{Setting} & \rotatebox{0}{Core Values} & \rotatebox{0}{Happiness} & \rotatebox{0}{Trust} & \rotatebox{0}{Econ.Int} & \rotatebox{0}{Security} & \rotatebox{0}{Tech} & \rotatebox{0}{Moral\&Rel} & \rotatebox{0}{Pol.Eng} & \rotatebox{0}{Demo} & \rotatebox{0}{\textbf{Overall}} \\
\midrule
\multirow{3}{*}{GPT-3.5-Turbo} & Full Info & 0.257 & 0.287 & 0.414 & 0.269 & 0.352 & 0.176 & 0.319 & 0.250 & 0.235 & 0.298 \\        
  & RAG & 0.448 & 0.224 & \textbf{0.508} & 0.311 & 0.447 & \textbf{0.249} & 0.502 & 0.343 & 0.307 & 0.404 \\
\rowcolor{ValueSimcolor}   & \textit{\textbf{ValueSim}} & \textbf{0.520} & \textbf{0.526} & 0.476 & \textbf{0.339} & \textbf{0.474} & 0.173 & \textbf{0.507} & \textbf{0.369} & \textbf{0.447} & \textbf{0.452} \\
\midrule
\multirow{3}{*}{Llama-3.1-8B} & Full Info & 0.289 & 0.137 & 0.439 & \textbf{0.299} & 0.383 & 0.145 & 0.222 & 0.290 & 0.252 & 0.305 \\
  & RAG & 0.345 & 0.133 & 0.433 & 0.245 & 0.361 & 0.177 & 0.435 & 0.276 & 0.276 & 0.337 \\
\rowcolor{ValueSimcolor}   & \textit{\textbf{ValueSim}} & \textbf{0.414} & \textbf{0.372} & \textbf{0.497} & 0.297 & \textbf{0.407} & \textbf{0.196} & \textbf{0.503} & \textbf{0.357} & \textbf{0.464} & \textbf{0.419} \\
\midrule
\multirow{3}{*}{Qwen-2.5-7B} & Full Info & 0.209 & 0.351 & 0.396 & 0.258 & 0.346 & \textbf{0.169} & 0.206 & 0.251 & 0.241 & 0.279 \\ 
  & RAG & 0.126 & 0.128 & 0.454 & 0.263 & 0.262 & 0.098 & 0.218 & 0.227 & 0.231 & 0.252 \\
\rowcolor{ValueSimcolor}   & \textit{\textbf{ValueSim}} & \textbf{0.472} & \textbf{0.496} & \textbf{0.521} & \textbf{0.300} & \textbf{0.438} & 0.127 & \textbf{0.300} & \textbf{0.358} & \textbf{0.425} & \textbf{0.414} \\
\midrule
\multirow{3}{*}{DeepSeek-V3} & Full Info & 0.495 & 0.196 & 0.540 & 0.276 & 0.468 & \textbf{0.309} & 0.489 & 0.429 & 0.304 & 0.437 \\ 
  & RAG & 0.466 & 0.347 & 0.565 & \textbf{0.363} & \textbf{0.545} & 0.282 & 0.510 & 0.438 & 0.353 & 0.465 \\
\rowcolor{ValueSimcolor}   & \textit{\textbf{ValueSim}} & \textbf{0.531} & \textbf{0.539} & \textbf{0.579} & 0.331 & 0.534 & 0.283 & \textbf{0.561} & \textbf{0.449} & \textbf{0.460} & \textbf{0.504} \\
\midrule
\multicolumn{2}{l}{{Chance Level}} & 0.255 & 0.174 & 0.213 & 0.147 & 0.242 & 0.091 & 0.151 & 0.177 & 0.138 & 0.194 \\
\bottomrule
\end{tabular}
\caption{Accuracy performance comparison (higher is better)}
\label{tab:accuracy_results}
\end{table*}

\begin{table*}[h]
\centering
\small
\setlength{\tabcolsep}{4pt}
\renewcommand{\arraystretch}{1.1}
\definecolor{ValueSimcolor}{rgb}{1.0,0.95,0.8}
\begin{tabular}{lcccc}
\toprule
\textbf{Setting} & \textbf{GPT-3.5-Turbo} & \textbf{Llama-3.1-8B} & \textbf{Qwen-2.5-7B} & \textbf{Deepseek-V3} \\
\midrule
\rowcolor{ValueSimcolor} ValueSim & \cellcolor{ValueSimcolor}\textbf{0.443} & \cellcolor{ValueSimcolor}\textbf{0.413} & \cellcolor{ValueSimcolor}\textbf{0.439} & \cellcolor{ValueSimcolor}\textbf{0.511} \\
ValueSim w/o CAB module & 0.375 & 0.405 & 0.407 & 0.456 \\
ValueSim w/o story module & 0.398 & 0.423 & 0.367 & 0.519 \\
\bottomrule
\end{tabular}
\caption{Accuracy comparison of different model variants across language models.}
\label{tab:accuracy_comparison}
\end{table*}

% Appendix C
\clearpage
\section{ValueSim Prompt Flow}
\label{app:ValueSim_prompt}

The ValueSim (Simulated Value-Based Generation) framework operates in a three-phase process designed to generate psychologically realistic user simulations aligned with specific value profiles. The framework consists of the following components:

\begin{enumerate}
    \item \textbf{Backstory Generation}: First, a comprehensive backstory is generated based on the user profile, creating a natural narrative that incorporates all demographic and value-related information.
    
    \item \textbf{Multi-dimensional Analysis}: Three parallel modules—cognitive, affective, and behavioral—analyze the question from different psychological perspectives. This approach is inspired by psychological and neuroscience research on human decision-making processes, which often involve different and sometimes contradictory mental systems.
    
    \item \textbf{Integrated Response}: Finally, a coordinator module synthesizes the analyses from the three perspectives to generate a cohesive final response.
\end{enumerate}

Below we provide the detailed prompts used in each component of our framework.

\subsection{Backstory Generation Module}
\label{app:backstory_module}

Unlike typical user simulations that rely on structured profiles, ValueSim transforms structured data into natural narratives that capture the user's background, beliefs, and values in a coherent storytelling format.

\begin{promptbox}[title=Backstory Generation Prompt]
\textbf{You are a background story writer, and you need to craft a comprehensive backstory for a person based on the information provided below.}

\textbf{IMPORTANT INSTRUCTIONS:}

1. Please rearrange and reorganize the sequence of this information to ensure it forms a coherent backstory.

2. YOU MUST INCLUDE EVERY SINGLE DATA POINT from the original information - no exceptions.

3. Do not summarize or generalize multiple data points - maintain the specific values, numbers, and exact responses.

4. Each information point in the data consists of a question, possible options, and the person's actual answer.

5. Focus primarily on the person's actual responses when creating the backstory.

6. Use second-person format throughout (e.g., "You believe..." "You were born in...").

7. Group related information together for coherence, but never at the expense of omitting details.

8. Format the backstory in clear paragraphs focusing on different aspects (demographics, beliefs, political views, etc.).

9. If the backstory becomes lengthy, that is acceptable - completeness is more important than brevity.

10. Please directly output the final backstory without returning any unnecessary content or explanations.

\textbf{Review your work carefully before submitting to ensure NO INFORMATION HAS BEEN OMITTED.}

\textbf{This person's Information:}\\
\{profile\_text\}
\end{promptbox}

\clearpage
\subsection{Multi-dimensional Analysis Modules}
\label{app:analysis_modules}

The multi-dimensional analysis phase employs three parallel modules that analyze the question from different psychological perspectives. These modules may produce different or even contradictory results, reflecting the complexity of human decision-making processes.

\begin{promptbox}[title=Cognitive Module]
\textbf{Please follow the Tutorial to analyze the User Profile below and answer the Question as if you were this person.}

\textbf{Tutorial:}\\
Consider these cognitive dimensions to understand this user:

- How does this user typically gather and prioritize information?

- What reasoning approaches do they seem to prefer?

- How might their beliefs and worldview frame this situation?

- Which factors would they likely weigh most heavily when deciding?

- What thinking patterns or cognitive tendencies might influence them?

\textbf{User Profile:}\\
\{backstory\}

\textbf{Question:}\\
\{question\_text\_with\_options\}

\textbf{Please format your response exactly as follows:}\\
Answer: [option number]\\
Analysis: [your reasoning for why this user would choose this option]
\end{promptbox}

\begin{promptbox}[title=Affective Module]
\textbf{Please follow the Tutorial to analyze the User Profile below and answer the Question as if you were this person.}

\textbf{Tutorial:}\\
Consider these affective dimensions to understand this user:

- What affective patterns and regulation styles characterize them?

- Which values and principles seem to guide their judgments?

- What affective needs or motivations might be activated here?

- How might they feel about the different possible outcomes?

- In what ways do their relationships and identity influence their feelings?

\textbf{User Profile:}\\
\{backstory\}

\textbf{Question:}\\
\{question\_text\_with\_options\}

\textbf{Please format your response exactly as follows:}\\
Answer: [option number]\\
Analysis: [your reasoning for why this user would choose this option]
\end{promptbox}

\begin{promptbox}[title=Behavioral Module]
\textbf{Please follow the Tutorial to analyze the User Profile below and answer the Question as if you were this person.}

\textbf{Tutorial:}\\
Consider these behavioral dimensions to understand this user:

- What behavioral tendencies and habits appear in their profile?

- How might their environment and social context influence their actions?

- Which capabilities and limitations might shape their behavioral choices?

- In what ways might past experiences guide their current decisions?

- How might they typically implement their decisions in practice?

\textbf{User Profile:}\\
\{backstory\}

\textbf{Question:}\\
\{question\_text\_with\_options\}

\textbf{Please format your response exactly as follows:}\\
Answer: [option number]\\
Analysis: [your reasoning for why this user would choose this option]
\end{promptbox}

\vspace{2cm}
\subsection{Coordinator Module}
\label{app:coordinator_module}

The coordinator module synthesizes the potentially divergent analyses from the three psychological perspectives to produce a final, integrated response that captures the complexity of human decision-making.

\begin{promptbox}[title=Coordinator Module]
\textbf{You are a coordinator in a user simulation system, and you need to synthesize analyses from three different perspectives to make a final decision.}

\textbf{Question:} \{question\_text\}\\
\textbf{Options:} \{options\_text\}

\textbf{Cognitive perspective answer:} \{cognitive\_data['answer']\}\\
\textbf{Cognitive perspective analysis:} \{cognitive\_data['analysis']\}

\textbf{Emotional perspective answer:} \{affective\_data['answer']\}\\
\textbf{Emotional perspective analysis:} \{affective\_data['analysis']\}

\textbf{Behavioral perspective answer:} \{behavioral\_data['answer']\}\\
\textbf{Behavioral perspective analysis:} \{behavioral\_data['analysis']\}

\textbf{Consider:}

• How their thoughts, feelings, and behavioral tendencies might interact in this situation

• Which aspects of their psychology seem most influential here

• Where their different perspectives align or create tension

\textbf{Format your response exactly as follows:}\\
Answer: [option number]\\
Analysis: [your reasoning for this decision]
\end{promptbox}

% Appendix D
\clearpage
\section{Backstory Examples}
\label{app:backstory_examples}

This section presents examples of backstories generated by the ValueSim framework using DeepSeek-V3 as the underlying language model. These examples were randomly selected from our test set and are presented in their entirety to demonstrate the richness and coherence of the narratives produced by our approach.

\subsection{Adult from Hungary (User ID: 3479)}
\label{app:user3479}

\begin{tcolorbox}[
    enhanced,
    breakable,
    %break at=9.5in, % This can be adjusted based on your page geometry
    colback=gray!5,
    colframe=gray!30,
    boxrule=0.5pt,
    arc=10pt,
    left=15pt,
    right=15pt,
    top=15pt,
    bottom=15pt,
    fontupper=\normalfont\small\fontfamily{ptm}\selectfont
]

\noindent\textbf{Backstory of User 3479}

You were born in 1973 in Hungary, where your mother was also born. Your father was born in this country as well. Your mother completed upper secondary education, while your father completed lower secondary education and belonged to the skilled worker group (e.g., foreman, motor mechanic, printer, seamstress, tool and die maker, electrician). You are currently 45 years old and living together as married. Your spouse has completed post-secondary non-tertiary education and is or was a full-time employee (30 hours a week or more). You are not the chief wage earner in your household, and you work or worked for a government or public institution, employed part-time (less than 30 hours a week).

You belong to the middle income group in your country and are moderately satisfied with the financial situation of your household (level 7 on a scale of 1-10). Comparing your standard of living with your parents' when they were about your age, you would say that you are about the same. Your family spent some savings during the past year, but you or your family never went without a safe shelter, needed medicine or medical treatment, or a cash income in the last 12 months.

You have a Master's degree or equivalent and belong to the professional and technical group (e.g., doctor, teacher, engineer, artist, accountant, nurse). You are a member of a professional association, an environmental organization, and an art, music, or educational organization, though you are not active in any of them. You are not a member of any women's group, self-help or mutual aid group, sport or recreational organization, consumer organization, humanitarian or charitable organization, church or religious organization, or labor union--though you are actively involved in a labor union.

Family is very important in your life, and you trust your family completely. You disagree that it is a duty towards society to have children, but you consider responsibility, imagination, tolerance and respect for other people, and determination and perseverance important qualities for children to learn. You do not consider independence, hard work, religious faith, obedience, thrift (saving money and things), or unselfishness important qualities for children.

You disagree that work is a duty towards society and that work should always come first, even if it means less spare time. However, work is rather important in your life, while leisure time is also rather important. You strongly disagree that on the whole, men make better business executives or political leaders than women do and that a university education is more important for a boy than for a girl. You agree that being a housewife is just as fulfilling as working for pay and that homosexual couples are as good parents as other couples.

You consider yourself not a religious person, though you believe in God, heaven, and hell. You pray once a year and attend religious services less often than once a year. God is neither important nor unimportant in your life (level 5 on a scale of 1-10). You believe the basic meaning of religion is to do good to other people instead of to follow religious norms and ceremonies, and to make sense of life in this world rather than to make sense of life after death. You disagree that your religion is the only acceptable one.

You place your political views at position 5 on the left-right scale (center position) and are not very interested in politics, which is not very important in your life. You have no confidence in political parties at all and not very much confidence in the government, parliament, elections, banks, major companies, television, courts, churches, charitable or humanitarian organizations, labor unions, or the World Trade Organization. However, you have quite a lot of confidence in the civil service, the International Criminal Court, the World Health Organization, the armed forces, the police, universities, women's organizations, and the International Monetary Fund.

You believe your country is extremely democratic in how it is being governed today (level 9 on a scale of 1-10) and that living in a country governed democratically is of extremely high importance to you (level 9 on a scale of 1-10). You believe international organizations should largely prioritize being democratic over being effective (level 8 on a scale of 1-10). You believe that people choosing their leaders in free elections is an absolutely essential characteristic of democracy (level 10 on a scale of 1-10), as is women having the same rights as men (level 10). However, you believe people receiving state aid for unemployment (level 3), civil rights protecting people from state oppression (level 4), governments taxing the rich and subsidizing the poor (level 4), people obeying their rulers (level 3), the army taking over when government is incompetent (level 1), religious authorities interpreting the laws (level 1), and the state making people's incomes equal (level 1) are not essential characteristics of democracy.

You believe opposition candidates are not often prevented from running in this country's elections, that election officials are fair fairly often, and that voters are offered a genuine choice fairly often, but you also believe voters are bribed fairly often and that rich people buy elections fairly often. You believe journalists do not often provide fair coverage of elections in this country and that most journalists and media people are involved in corruption, along with most business executives, state authorities, and local authorities, though you think few civil service providers are involved in corruption. You believe there is substantial corruption in your country (level 8 on a scale of 1-10) and that there is a considerable risk of being held accountable for bribery (level 7).

You believe in gradual societal improvement through reforms and that maintaining order in the nation is most important for the country, followed by a high level of economic growth as the most important goal for the next ten years, people having more say about how things are done at their jobs and in their communities as the second most important goal, and giving people more say in important government decisions as the second most important. You moderately believe in greater incentives for individual effort, with limited support for income equality (level 8) and that people should take more responsibility to provide for themaelves, with limited emphasis on government responsibility (level 8). You somewhat believe in competition, with some concerns about its harm (level 4) and that hard work usually brings a better life, though luck and connections also matter (level 4).

You believe somewhat equally in both private and government ownership, with a slight preference for government ownership (level 6). You feel it would be fairly bad to have experts, not government, make decisions for the country and very bad to have a strong leader who bypasses parliament and elections or to have the army rule. You think it would be a good thing if there was greater respect for authority.

You agree that immigration leads to social conflict and increases the crime rate, but you find it hard to say whether immigration increases unemployment, strengthens cultural diversity, offers a better life to people from poor countries, or increases the risks of terrorism. You believe immigrants have neither a good nor bad impact on your country's development. You do not mind having immigrants/foreign workers, people of a different race, people who speak a different language, homosexuals, or people who have AIDS as neighbors, but you would not like to have heavy drinkers or drug addicts as neighbors.

You trust people you know personally somewhat, people of another nationality somewhat, and your neighborhood somewhat, but you do not trust people you meet for the first time very much. You feel close to your country, county/region/district, and village/town/city, but not close to your continent or the world at all.

You or your family never felt unsafe from crime in your home in the last 12 months, and no one in your family has been a victim of crime in the past year. Robberies, drug sales, street violence, and sexual harassment do not occur frequently (or at all) in your neighborhood, and alcohol consumption in the streets does not occur frequently. You have avoided going out at night for security reasons but have not carried a weapon for security reasons. You feel quite secure these days and, if forced to choose, would consider security more important than freedom.

You believe violence against other people is never justified (level 1), including a man beating his wife (level 1), parents beating children (level 1), and terrorism as a political, ideological, or religious means (level 1). You believe suicide is almost never justified (level 2), claiming government benefits you are not entitled to is rarely justified (level 3), someone accepting a bribe is rarely justified (level 3), avoiding a fare on public transport is usually not justified (level 4), cheating on taxes is somewhat not justified (level 5), prostitution is somewhat not justified (level 5), homosexuality is somewhat not justified (level 5), the death penalty is somewhat not justified (level 5), euthanasia is somewhat justified (level 6), abortion is somewhat justified (level 6), divorce is sometimes justified (level 7), and having casual sex (level 8) and sex before marriage (level 8) are often justified.

You moderately disagree that we depend too much on science and not enough on faith (level 3) and that science breaks down people's ideas of right and wrong (level 3). You believe the world is moderately better off because of science and technology (level 7) but neither agree nor disagree that science and technology make our lives healthier, easier, and more comfortable (level 5) or that they will create more opportunities for the next generation (level 5). You don't mind if there was more emphasis on technology development.

You always vote in national and local elections and might encourage others to vote or take political action, though you have not yet organized political activities online or searched for political information online. You have signed a petition and an electronic petition before, have contacted a government official, and have joined strikes before. You have donated to a group or campaign before.

You obtain information from TV news daily, the Internet daily, talking with friends or colleagues daily, and radio news weekly. You do not have very much confidence in television.

You describe your state of health these days as good and are very happy overall. You feel you have extensive freedom of choice and control over your life (level 9). You are not worried much about a war involving your country, a civil war, losing your job, or not being able to give your children a good education.

You believe in a democratic, orderly society with economic growth and individual responsibility, though you are critical of corruption and inequality. You value security, gradual reform, and personal freedoms, while maintaining a moderate, balanced outlook on most issues. Your life is shaped by family, work, and a cautious but hopeful view of the world.

\end{tcolorbox}

% Appendix E
\clearpage
\section{Baseline Prompts}
\label{app:baseline_prompts}

For comparison purposes, we implemented two baseline approaches to simulate human responses based on value profiles. Below are the exact prompts used for each baseline.

\subsection{Direct Profile Approach}
\label{app:direct_profile}

This baseline represents the conventional approach where the original structured profile is directly provided to the model without any narrative transformation or multi-perspective analysis.

\begin{promptbox}[title=Direct Profile Baseline Prompt]
\textbf{Question:} \{question\_text\_with\_options\}

\textbf{User profile:} \{original\_profile\_text\}

\textbf{Consider both the question context and the user's background when formulating your response. Aim for a balanced perspective that respects accuracy while reflecting the user's viewpoint.}

\textbf{Answer format:} 'option you selected'
\end{promptbox}

\subsection{Retrieval-Augmented Approach}
\label{app:rag_approach}

This baseline employs a retrieval-augmented generation approach, where only the most relevant portions of the user profile (typically the top three most relevant information points) are provided to the model.

\begin{promptbox}[title=Retrieval-Augmented Baseline Prompt]
\textbf{Question:} \{question\_text\_with\_options\}

\textbf{Relevant user information:}
\{retrieved\_profile\_segments\}

\textbf{Based ONLY on the relevant user information provided above, answer the question. Consider both the question context and the user's background from the provided relevant information. Aim for a balanced perspective that respects accuracy while reflecting the user's viewpoint.}

\textbf{Answer format:} 'option you selected'
\end{promptbox}

% Appendix F
\clearpage
\section{World Values Survey Dataset Details}
\label{app:wvs_dataset}

This section details how we structured and organized value categories from the World Values Survey (WVS) dataset for use in our ValueSim framework. We first present the original WVS structure and then explain our domain-specific reorganization approach.

\subsection{Original WVS Dataset Structure}

The World Values Survey Wave 7 provides a comprehensive collection of cross-cultural data on human values, covering surveys from 66 countries/territories. The questionnaire consists of approximately 290 questions organized into 14 thematic sections as shown in Table \ref{tab:original_wvs}.

\begin{table}[h]
\centering
\caption{Original Value Categories and Question Mappings from WVS}
\label{tab:original_wvs}
\begin{tabular}{p{7cm}p{5cm}}
\hline
\textbf{Original WVS Category} & \textbf{Question Numbers} \\
\hline
Social Values, Attitudes \& Stereotypes & Q1-Q45 \\
\hline
Happiness And Well-Being & Q46-Q56 \\
\hline
Social Capital, Trust \& Organizational Membership & Q57-Q105 \\
\hline
Economic Values & Q106-Q111 \\
\hline
Corruption & Q112-Q120 \\
\hline
Migration & Q121-Q130 \\
\hline
Security & Q131-Q151 \\
\hline
Postmaterialist Index & Q152-Q157 \\
\hline
Science And Technology & Q158-Q163 \\
\hline
Religious Values & Q164-Q175 \\
\hline
Ethical Values And Norms & Q176-Q198 \\
\hline
Political Interest And Participation & Q199-Q234 \\
\hline
Political Culture And Regimes & Q235-Q259 \\
\hline
Demographics & Q260-Q290 \\
\hline
\end{tabular}
\end{table}

\subsection{Thematic Reorganization for ValueSim}

To optimize the value representation in our ValueSim framework, we developed a more consolidated categorization system that groups semantically related value dimensions. This reorganization creates more coherent thematic units while preserving the comprehensive coverage of the original WVS structure.

\subsubsection{Rationale for Category Reorganization}

Our thematic reorganization was guided by several principles:

\begin{itemize}
    \item \textbf{Conceptual coherence:} We merged categories that measure closely related value constructs, such as combining religious values with ethical norms due to their significant conceptual overlap in many cultural contexts.
    
    \item \textbf{Analytical practicality:} Consolidating the original 14 categories into 9 broader dimensions creates a more manageable taxonomy for analysis and visualization, while still capturing the multidimensional nature of human values.
    
    \item \textbf{Value interdependencies:} Our reorganization acknowledges how certain value domains naturally cluster together, such as security concerns and migration attitudes, which often share underlying perspectives on social boundaries and perceived threats.
    
    \item \textbf{Interpretability:} The consolidated categories provide more robust constructs for analyzing similarities and differences in value patterns, enhancing the interpretability of cross-cultural comparisons.
\end{itemize}

Table \ref{tab:reorganized_wvs} presents our reorganized value categories, their abbreviations used in visualizations, and their mapping to the original WVS categories.

\begin{table}[h]
\centering
\caption{Reorganized Value Categories for ValueSim Framework}
\label{tab:reorganized_wvs}
\begin{tabular}{p{5cm}p{2cm}p{7cm}}
\hline
\textbf{Reorganized Category} & \textbf{Abbreviation} & \textbf{Original WVS Categories} \\
\hline
Core Value Orientations & Core & Social Values, Attitudes \& Stereotypes; Postmaterialist Index \\
\hline
Happiness and Well-being & Hap. & Happiness And Well-Being \\
\hline
Social Capital, Trust and Organizational Membership & Trust & Social Capital, Trust \& Organizational Membership \\
\hline
Economic Integrity & Econ.Int & Economic Values; Corruption \\
\hline
Security-Migration Nexus & Security & Security; Migration \\
\hline
Science and Technology & Tech & Science And Technology \\
\hline
Moral-Religious Framework & Mo.\&Rel. & Religious Values; Ethical Values And Norms \\
\hline
Political Engagement & Pol.Eng & Political Interest And Participation; Political Culture And Regimes \\
\hline
Demographics & Demo & Demographics \\
\hline
\end{tabular}
\end{table}

\subsubsection{Category Integration Explanations}

Our thematic reorganization reflects meaningful connections between value dimensions:

\begin{itemize}
    \item \textbf{Core Value Orientations} integrates general social values with the postmaterialist index, as both address fundamental value priorities and social orientations that form the foundation of a person's worldview.
    
    \item \textbf{Economic Integrity} combines economic value orientations with attitudes toward corruption, acknowledging the interrelationship between economic systems and institutional integrity in shaping perspectives on fair resource allocation.
    
    \item \textbf{Security-Migration Nexus} recognizes the conceptual link between personal/national security concerns and attitudes toward migration and foreigners, which often reflect similar underlying orientations toward societal boundaries and perceived external influences.
    
    \item \textbf{Moral-Religious Framework} acknowledges the substantial overlap between religious values and ethical norms in many cultural contexts, where religious beliefs often inform moral judgments on various social issues.
    
    \item \textbf{Political Engagement} brings together political interest/participation with broader political culture attitudes, creating a more comprehensive representation of how individuals engage with and conceptualize political systems.
\end{itemize}

In terms of question mapping, each reorganized category encompasses all question numbers from its constituent original categories. For instance, the "Core Value Orientations" category includes questions Q1-Q6, Q27-Q45 (from Social Values) and Q152-Q157 (from Postmaterialist Index).

This reorganized categorization system provides a more streamlined yet comprehensive framework for analyzing human values within our ValueSim approach, enabling more intuitive interpretation of value patterns while maintaining the rich empirical foundation of the original WVS dataset.

\end{document}